\documentclass{article}

\usepackage{graphicx} 
\usepackage{subfigure}

\usepackage{algorithm}
\usepackage{algorithmic}
\usepackage{appendix}

\usepackage{times}
\usepackage{epsfig}
\usepackage{graphicx}
\usepackage{amsmath}
\usepackage{amssymb}
\usepackage{graphics}
\usepackage{graphicx}
\usepackage[]{graphicx} 
\usepackage{epsfig} 
\usepackage{subfigure}
\usepackage{stmaryrd}
\usepackage[mathscr]{eucal}
\usepackage{lineno}
\usepackage{color}
\usepackage{filecontents}
\usepackage{subfigure}
\usepackage{multirow}
\usepackage{amsmath}
\usepackage{amssymb}
\usepackage{filecontents}
\usepackage{verbatim} 
\usepackage{tabularx}
\usepackage{lineno}
\usepackage{setspace}

\usepackage{hyperref}

\input{macros}

\def\A{{\bf A}}

\def\B{{\bf B}}

\def\C{{\bf C}}

\def\D{{\bf D}}
\def\d{{\bf d}}

\def\F{{\bf F}}

\def\K{{\bf K}}
\def\H{{\bf H}}

\def\I{{\bf I}}
\def\N{{\bf N}}
\def\R{{\bf R}}
\def\X{{\bf X}}
\def\Y{{\bf Y}}

\def\S{{\bf S}}
\def\x{{\bf x}}
\def\y{{\bf y}}

\def\Z{{\bf Z}}
\def\M{{\bf M}}
\def\m{{\bf m}}
\def\n{{\bf n}}

\def\W{{\bf W}}
\def\w{{\bf w}}
\def\0{{\bf 0}}
\def\1{{\bf 1}}

\def\RB{{\mathbb R}}

\def\alp{\mbox{\boldmath$\alpha$\unboldmath}}

\def\ie{\emph{i.e. }}

\def\argmin{\mathop{\rm argmin}}

\def\etal{{\em et al.\/}\,}

\title{Online Discriminative Dictionary Learning for Image Classification Based on Block-Coordinate Descent Method}

\author{
Shu Kong and Donghui Wang\\
\texttt{\{aimerykong, dhwang\}@zju.edu.cn}\\
College of Computer Science and Technology, Zhejiang University\\
Hangzhou, China
}

\date{\today}

\begin{document}
\maketitle

\begin{abstract}
Previous researches have demonstrated that the framework of dictionary learning with sparse coding, in which signals are decomposed as linear combinations of a few atoms of a learned dictionary, is well adept to reconstruction issues.
This framework has also been used for discrimination tasks such as image classification. 
To achieve better performances of classification, experts develop several methods to learn a discriminative dictionary in a supervised manner. 
However, another issue is that when the data become extremely large in scale, these methods will be no longer effective as they are all batch-oriented approaches.
For this reason, we propose a novel online algorithm for discriminative dictionary learning, dubbed \textbf{ODDL} in this paper.
First, we introduce a linear classifier into the conventional dictionary learning formulation and derive a discriminative dictionary learning problem. 
Then, we exploit an online algorithm to solve the derived problem. 
Unlike the most existing approaches which update dictionary and classifier alternately via iteratively solving sub-problems, our approach directly explores them jointly. 
Meanwhile, it can largely shorten the runtime for training and is also particularly suitable for large-scale classification issues. 
To evaluate the performance of the proposed ODDL approach in image recognition, we conduct some experiments on three well-known benchmarks, and the experimental results demonstrate ODDL is fairly promising for image classification tasks.
\end{abstract}

\section{Introduction}
\label{intro}
Dictionary learning with sparse coding, which decompose signals as linear combinations of a few atoms from some basis or dictionary, have drawn extensive attentions in recent years.
Researchers have demonstrated that this framework can achieve state-of-the-art performances in image processing tasks such as image denoising~\cite{denosing_CVPR2006}, face recognition~\cite{Pham08,Zhang10}, etc. 
Given a signal $\x \in \RB^n$ and a fixed dictionary $\D \in \RB^{n \times k}$ which contains $k$ atoms, we say that $\x$ admits a sparse representation over $\D$, if we can find one sparse coefficient $\alp \in \RB^k$ which makes $\x \approx \D\alp$. As we know, predefined dictionaries, based on various types of wavelets~\cite{wavelet}, are not suitable for many vision applications such as appearance-based image classification, because the atoms of these dictionaries do not make use of the semantic prior of the given signals. However, the learned dictionaries can achieve more promising performances in various image processing tasks than that of the predefined ones~\cite{Mairal07,Ma2010}.

Several algorithms have been proposed for learning such dictionaries based on sparse representation recently. For example, K-SVD algorithm~\cite{K-SVD} is one such algorithm which learns an overcomplete dictionary from the training data. 
It updates the atoms in the dictionary one at a time, by fixing all the other atoms unchanged and finding a new one with its corresponding coefficients which minimize the mean square error (MSE). 
Researchers have shown that this algorithm can achieve outstanding performances in image compression and denosing~\cite{compression,imagedenoising}. 
However, K-SVD algorithm merely focuses on the reconstructive power of learned dictionary, thus it is intrinsically adapted for (image) discrimination or classification tasks. 
To address this problem and to make use of dictionary learning powerfulness, several methods have been proposed recently. 
For example, semi-supervised dictionaries~\cite{Pham08} are learned via updating the K-SVD dictionary based on results of a linear classifier iteratively. 
As well, by adding a linear classifier, another algorithm called discriminative K-SVD~\cite{Zhang10} is developed for image classification.
Moreover, to obtain the discriminative capability of the dictionary, a more sophisticated loss function called logistic loss function (softmax function for multiclass classification) is added to the classical dictionary formulation ~\cite{DDL,Mairal09}.

In addition, most recent methods for dictionary learning are iterative batch algorithms, which assess all the training samples at each iteration to minimize the objective function under sparse constraints. Therefore, another problem we may encounter is that when the training set becomes very large, these methods are no longer efficient. 
To overcome this bottleneck, an online algorithm for dictionary learning which applies block-coordinate descent method~\cite{online} has been proposed in the literature. 
However, this online dictionary learning method is still learning the reconstructive dictionary which can well represent the signals, but is not adapted for classification. 
Marial \etal attempt to address this issue by task-driven dictionary learning~\cite{task-dl} where supervised dictionaries are learned via a stochastic gradient descent algorithm.

To overcome the above two problems, \ie lacking discriminative power in the reconstructive dictionary and the issue caused by large-scale training set, we propose a novel online discriminative formulation for learning the discriminative dictionaries in a online manner.
We name our approach \textbf{ODDL} in this paper.
In our work, we first incorporate label information into the dictionary learning stage by adopting a linear classifier, and then formulate a supervised dictionary learning problem. 
To solve this problem, we propose a corresponding online algorithm, in which we apply the block-coordinate descent method to train the dictionary and classifier simultaneously. 
Unlike most recent methods which update the dictionary and classifier alternately via iteratively exploring the solution of sub-problems, it directly learns the dictionary and classifier jointly. Finally, we carry out some experiments on three well-known benchmarks to demonstrate the effectiveness of our proposed method, and the experimental results show the proposed ODDL method is fairly competitive for classification tasks.

In summary, the main contributions of this paper include the following:
\begin{itemize}
\item We propose a novel online algorithm with the numerical solution to learn a discriminative dictionary. It enables online framework and learning discriminative dictionary to merge into one framework. In other words, our proposed approach can efficiently and effectively derive the discriminative dictionary, meanwhile it overcomes large scale classification problem.
\item By analysis, we see our algorithm can update the classifier simultaneously with the update of the dictionary when a new training sample comes. By this way, computational cost can be significantly reduced.
\item As shown experimentally, our approach achieves encouraging performance compared with some other dictionary learning approaches.
\item Interestingly, we suggest a novel, efficient and effective dictionary construction scheme for face recognition. This scheme shows lights on face recognition experimentally.
\end{itemize}

The paper is organized as follows. Section 2 introduces the basic formulation of dictionary learning and sparse representation for classification. Then our proposed approach is presented in Section 3, followed by the experimental results demonstrated in Section 4. Finally, we conclude our paper in Section 5.

\section{Related Work on Dictionary Learning Methods}
\label{sec:2}

\begin{figure}[!t]
    \centering
    \includegraphics[width=0.8\linewidth]{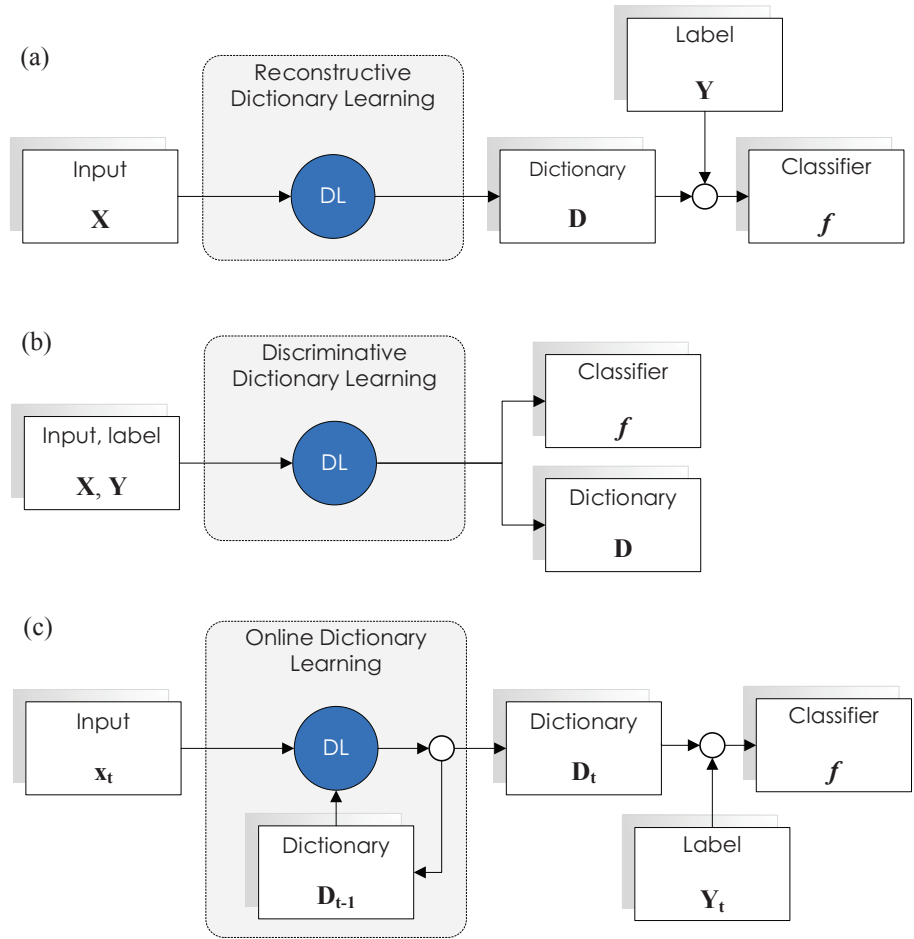}
    \caption{ Flows of three different dictionary learning schemes. From the top to bottom, the schematic illustration of dictionary learning methods are reconstructive (a), discriminative (b) and online (c).}
    \label{fig:flowChart}
\end{figure}

Recent researches have demonstrated that natural signals such as images can admit sparse representations of some redundant basis\footnote{Here the term ``basis'' is loosely used, since the dictionary can be overcomplete and, even in the case of just complete, there is no guarantee of independence between the atoms.} (also called dictionary). This phenomenon can explain the feasibility that image classification can be done by sparse representation with an overcomplete dictionary learned from the training images. In this section we briefly review three dictionary learning schemes which are closely relevant to our proposed method. 
Fig.~\ref{fig:flowChart} illustrates the flows of the three dictionary learning schemes with a classifier training process.

\subsection{Reconstructive Dictionary Learning for Classification}

In classical sparse coding problems, consider a signal
$\x \in \RB^{n}$ and a dictionary $\D = [\d_1,...,\d_k] \in \RB^{n \times k}$.
Under the assumption that a natural signal can be approximately represented
by a linear combination of a few selected atoms from the dictionary, $\x$ then can
be represented by $\D\alp$ for some sparse coefficient vectors $\alp \in \RB^{k}$. To find the sparse representation of $\x$ is equivalent to the following optimization problem:
\begin{equation}
\mathop{\min}\limits_{\alp \in \RB^k} \mathop{\left\|\x-\D\alp\right\|}\nolimits_2^2,\textrm { \emph{s.t.} }\mathop{\left\|\alp\right\|}\nolimits_p \le L
\label{equa:1}
\end{equation}
where $p$ is $0$ or $1$. The $\ell^0$ pseudo norm sparse coding is an NP-hard problem~\cite{NP-hard} and several greedy algorithms~\cite{MP,OMP} have been proposed to approximate the solution. The $\ell^1$ formulation of sparse coding is the well-known Lasso~\cite{Lasso} or Basic Pursuit~\cite{BP} problem and can be effectively solved by algorithms such as LARS~\cite{LARS}.

Eq.~\ref{equa:1} is the classical reconstructive dictionary learning problem, in which overlapping patches instead of the whole images are sparsely decomposed as a result of the natural images are usually very large. 
For an image $\I$, suppose there are $M$ overlapping patches ${\{\x_i\}}_{i=1}^M \in \RB^n$ from image $\I$. 
Then the dictionary $\D \in \RB^{n \times k}$ is learned via alternatively solving the following optimization over $\D$ and $\A$:
\begin{equation}
\{\D,\A\} = 
\argmin\limits_{
\substack{ \D \in \RB^{n \times k} \\
\A \in \RB^{k \times M}
}} \mathop{\sum}\limits_{i=1}^M\mathop{\left\|\x_i-\D\alp_i\right\|}\nolimits_2^2,\textrm { \emph{ s.t. } }\mathop{\left\|\alp_i\right\|}\nolimits_p \le L \textrm{ for } i=1,\dots,M,
\label{equa:2}
\end{equation}
where $\A = [\alp_1,...,\alp_M]\in \RB^{k \times M}$ is the coefficient matrix, $\x_i$ is the $i^{th}$ patch of image $\I$ written as a column vector, $\alp_i$ is the corresponding sparse code. Several algorithms have been proposed to solving this dictionary learning problem, such as \cite{K-SVD} and \cite{MOD}.

Given $C$ sets of signals $P_i, i = 1,2,...,C,$ which belong to $C$ different classes. The training stage for classification based on sparse representations is composed of two independent parts:  dictionary learning and classifier learning. First, a dictionary $\D$ of $C$ classes is learned according to (\ref{equa:2}). Then, the classifier is trained via solving the following optimization problem:
\begin{equation}
\mathop{\min}\limits_{\W}f(\Y,\W,\A)+\lambda{\left\|\W\right\|}_F^2
\label{equa:4}
\end{equation}
where $\Y$ is the label matrix of the training pathes, $\A$ is the coefficient matrix computed on the learned dictionary $\D$, and $f$ is a loss function.
However, this dictionary learning scheme has two main drawbacks, easily shown in Fig.~\ref{fig:flowChart} (a):

1. The dictionary training and classifier training are two independent stages. Thus, the learned dictionary cannot capture the most discriminative cues that are helpful for classification.

2. Practically, to improve the representative capacity of the dictionary, we often exploit large-scale training samples to obtain a powerful dictionary in representation. But this action actually will fail to learn an effective dictionary, due to the large-scale dataset problem.

\subsection{Discriminative Dictionary Learning for Classification}

Researchers have already made some efforts to overcome the first drawback mentioned in previous subsection that the learned dictionaries lack discrimination power for classification. 
In~\cite{DDL,Mairal09}, a discriminative term is introduced to combine the classifier learning process with dictionary learning, and the final objective function is:
\begin{equation}
\begin{split}
&\mathop{\min}\limits_{\A,\D,\W}\mathop{\sum}\limits_{i}
\bigg\{
\Vert \x_i - \D\alp\Vert_2^2 +
C(y_i \cdot f(\alp_i,\W))
\bigg\}
+\lambda_1\mathop{\left\|\W\right\|}\nolimits_F^2, \\
&\textrm{ s.t. } \alp_i \le L \textrm{ for } i=1,\dots,M,
\end{split}
\label{equa:3}
\end{equation}
where $\W$ is the classifier parameter, $y_i = 1$ or $-1$ is the label of patch $\x_i$, and $C$ is a logistic loss function, $C(x) = log(1+e^{-x})$.
In addition, in \cite{Pham08} and \cite{Zhang10}, a simpler term which is a linear classifier is considered for the discriminative power:
\begin{equation}
\begin{split}
&\mathop{\min}\limits_{\A,\D,\W}\mathop{\sum}\limits_{i}
\bigg\{
\Vert \x_i - \D\alp\Vert_2^2 +
\mathop{\left\|\y_i-\W\alp_i-\bf{b}\right\|}\nolimits_2^2
\bigg\}
+\lambda_1{\left\|\W\right\|}_F^2, \\
&\textrm{ s.t. } \alp_i \le L \textrm{ for } i=1,\dots,M,
\end{split}
\label{equa:4}
\end{equation}
where $\W$ and $\bf{b}$ are the classifier parameters, $\y_i$ is the label vector of patch $\x_i$ in which the element associated with the class label is 1 and the others are 0. ${\left\|\cdot\right\|}_F$ denotes the Frobenius norm of a matrix $\X$, \ie ${\left\|\X\right\|}_F = ({\sum}_i{\sum}_j{x_{ij}}^2)^{1/2}$.
Without generalization, the intercept $\bf{b}$ can be omitted by normalize all the signals.

Dictionaries learned by these methods generally perform better in classification tasks than those learned in a reconstructive way. However, from Fig.~\ref{fig:flowChart} (b), we can see a fatal drawback of these methods is that, if a new and important training sample comes after the dictionary has been learned, we have to relearn the dictionary from scratch.
From another point of view, discriminant dictionary learning methods suffer from large-scale dataset problem.

\subsection{Online Dictionary Learning for Classification}
Large-scale training set is a reasonable extension from human beings in learning from experiences. 
But the aforementioned two dictionary learning schemes fail to handle large-scale dataset problem. 
For this reason, an online dictionary learning algorithm~\cite{online} turns up to an efficient dictionary learning paradigm for large-scale training set. 
Inspired by~\cite{Bottou07}, Mairal \etal use the expected objective function to replace the original empirical objective function, obtaining an novel dictionary learning problem:
\begin{equation}
\mathop{\min}\limits_\D\frac{1}{2}{\mathbb{E}}_{\x}[{\left\|\x-\D\alp^*\right\|}_2^2]
\label{equa:5}
\end{equation}
where $\alp^*$ denotes the sparse coefficients computed in the sparse coding stage. To solve the above problem, they propose an online algorithm which applies the block-gradient descent method for dictionary updating. However, one obvious drawback of this algorithm is that it also ignores the valuable label information which will enhance classification performance. Furthermore, from the flow of training process reflected in Fig.~\ref{fig:flowChart} (c), another critical defect can be easily seen that even though the dictionary can be efficiently learned in an online manner, the classifier must be relearned from scratch when a new training sample comes.

\section{Online Discriminative Dictionary Learning}
\label{sec:3}
\begin{figure}[!t]
    \centering
    \includegraphics[width=0.6\linewidth]{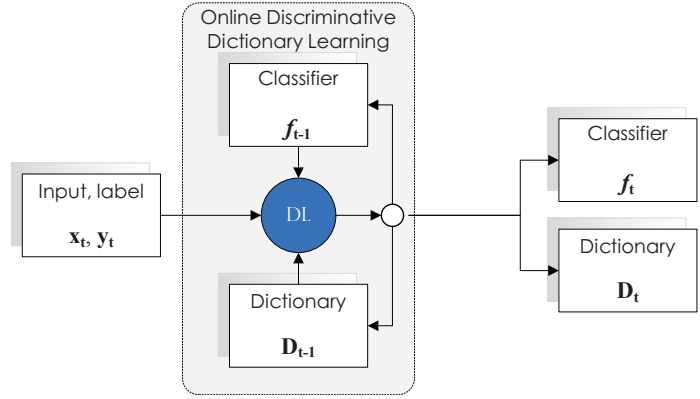}
    \caption{ The flow of our proposed dictionary learning algorithm.}
    \label{fig:ODDL}
\end{figure}

In the previous section, we review three dictionary learning schemes with their respective drawbacks. 
Now we derive our online discriminative dictionary learning (ODDL) to overcome the mentioned defects.
The schematic flow chart is demonstrated in Fig.~\ref{fig:ODDL}, from which we can see the obvious difference from the aforementioned three schemes.

\subsection{Proposed Formulation}

To overcome the issue lack of discriminative information for learned dictionary, we introduce an discriminative term to the original dictionary learning problem. In this paper, we consider the linear classifier for its simplicity. Adding the linear classifier, we obtain the following problem:
\begin{equation}
\mathop{\min}\limits_{\D,\W}\mathop{\left\|\X-\D\A\right\|}\nolimits_F^2 +\lambda_0\mathop{\left\|\Y-\W\A\right\|}\nolimits_F^2 + \lambda_1\mathop{\left\|\W\right\|}\nolimits_F^2,
\label{equa:8}
\end{equation}
where $\X$ is the patch matrix, $\Y$ is the label matrix of the training patches, ${\left\|\X-\D\A\right\|}_F^2$ is the reconstructive error term, ${\left\|\Y-\W\A\right\|}_F^2$ is the discriminative term, and $\lambda_0$ controls the trade-off between the reconstructive and discriminative terms.

Now we need to address another issue about the large-scale dataset problem, as Bottou \etal~\cite{Bottou07} say, the minimization of the empirical cost is not the focus of researchers, but instead the minimization of the expected cost:
\begin{equation}
\mathop{\min}\limits_{\D,\W}{\mathbb{E}}_{\x,\y}[{\left\|\x-\D\alp^*\right\|}_2^2 +\lambda_0{\left\|\y-\W\alp^*\right\|}_2^2] +\lambda_1{\left\|\W\right\|}_F^2
\label{equa:9}
\end{equation}
where the expectation is taken with respect to the joint distribution of $(\x,\y)$. In practice, to improve the representative power of learned dictionaries, a large amount of training data is always needed. For example, when applying dictionary for image processing tasks, the number of training patches can be up to several millions in a single image. In this case, we must exploit an efficient technique to solve this large-scale dataset problem and online learning is such a technique.

\subsection{Optimization}

In this subsection, we briefly introduce an online discriminative dictionary learning algorithm to solve the proposed formulation (\ref{equa:9}) in the previous subsection. As same as most existing dictionary learning algorithms, there are still two stages in our proposed algorithm.

\textbf{Sparse coding} The sparse coding problem (\ref{equa:1}) with learned dictionary $\D$ is an $\ell^p$ norm optimization problem, where $p$ is 0 or 1. Several algorithms have been proposed for solving this problem. In this paper, we choose the $\ell^0$ pseudo norm optimization problem as our sparse coding problem since in this formulation we can explicitly control the sparsity (nonzero elements) of the coefficients of the signals projected on the learned dictionary. This leads us to use the Orthogonal Matching Pursuit (OMP) algorithm~\cite{OMP}, a greedy algorithm which selects atoms with highest correlation to current orthogonal projected residual sequentially.

\textbf{Dictionary and classifier updating} This stage is markedly different from that of other discriminative dictionary learning approaches. In our proposed ODDL, we use the block-coordinate descent method for updating dictionary and classifier jointly, while the usual strategy in other algorithms consists of finding the approximate global solutions of dictionary and classifier via solving sub-problem iteratively.

Rewrite Eq.~\ref{equa:8} and we can derive a compact formulation as our objective function:
\begin{equation}
\mathop{\min}\limits_{\D,\W}{\mathbb{E}}_{\x,\y}[{\left\|{{\x \choose \sqrt{\lambda_0}\y}}-{\D \choose \sqrt{\lambda_0}\W}\alp\right\|}_2^2] +\lambda_1{\left\|\W\right\|}_F^2.
\end{equation}
Note that from a dictionary learning viewpoint, the ``dictionary'' ${\D \choose \sqrt{\lambda_0}\W}$, which represents the ``signal'' ${{\x \choose \sqrt{\lambda_0}\y}}$, is always assumed to be normalized column-wise in updating process, \ie the Euclidian length of columns in the ``dictionary'' is 1. 
Moreover, the real dictionary $\D$ we derive is also normalized, therefore, we can drop the regularization term ${\left\|\W\right\|}_F^2$ in the objective function. 
Thus, we derive the final function:
\begin{equation}
\mathop{\min}\limits_{\D,\W}{\mathbb{E}}_{\x,\y}[{\left\|{{\x \choose \sqrt{\lambda_0}\y}}-{\D \choose \sqrt{\lambda_0}\W}\alp\right\|}_2^2]
\label{equa:10}
\end{equation}
In our algorithm, there is an important assumption that the training set is composed of i.i.d. samples $(\x,\y)$ which admits a probability distribution $p(\x,\y)$. Using the same strategy in stochastic gradient descent, our algorithm draws one sample $(\x_t,\y_t)$ at each iteration, and computes the sparse code $\alp_t$ of $\x_t$ on the previous dictionary $\D_{t-1}$, then updates dictionary $\D_t$ and classifier parameter $\W_t$ simultaneously via solving the following problem
\begin{equation}
\mathop{\min}\limits_{\D,\W}\mathop{\sum}\limits_{i=1}^t{\left\|{{\x_i \choose \sqrt{\lambda_0}\y_i}}-{{\D \choose \sqrt{\lambda_0}\W}}\alp_i\right\|}_2^2
\label{equa:11}
\end{equation}
To address this problem, first we denote ${{\x_i \choose \sqrt{\lambda_0}\y_i}}$ as $\mathop{\widetilde{\x}}\nolimits_i$ and ${{\D \choose \sqrt{\lambda_0}\W}}$ as $\widetilde{\D}$. Then problem (\ref{equa:11}) can be rewritten as
\begin{equation}
\begin{array}{c}
\mathop{\min}\limits_{\D,\W}Tr({\widetilde{\D}}^T \widetilde{\D}\M_t) -
2Tr({\widetilde{\D}}^T \N_t),\\
 \textrm { \emph{ where } }
\M_t = \mathop{\sum}\limits_{i=1}^t\alp_i{\alp_i}^T, \N_t = \mathop{\sum}\limits_{i=1}^t\mathop{\widetilde{\x}}\nolimits_i{\alp_i}^T
\label{equa:12}
\end{array}
\end{equation}
Using the block-coordinate descent method, the $j$-th column of $\widetilde{\D}$ can be updated using
\begin{equation}
\mathop{\widetilde{\bf{d}}}\nolimits_j = \frac{1}{m_{jj}}(\n_j - \D\m_j)+\mathop{\widetilde{\bf{d}}}\nolimits_j
\label{equa:13}
\end{equation}
Then parting $\d_j$ and $\w_j$ off $\mathop{\widetilde{\bf{d}}}\nolimits_j$ we can update $\bf{d}_j$ and $\w_j$ by
\begin{equation}
\begin{array}{lcr}
\d_j & = & \d_j/{\left\|\d_j\right\|}_2\\
\w_j & = & \w_j/{\left\|\d_j\right\|}_2
\label{equa:14}
\end{array}
\end{equation}
The details of derivation are showed in Appendix.

\subsection{Algorithm}

The approach we propose in this paper is a block-coordinate descent algorithm, and the overall algorithm is summarized in Algorithm~\ref{alg:1}. 
In this algorithm, the i.i.d. samples $(\x_t,\y_t)$ are drawn from an unknown probability distribution $p(\x,\y)$ sequentially. However, since the distribution $p(\x,\y)$ is unknown, obtaining such i.i.d. samples may be very difficult. The common trick in online algorithms to obtain such i.i.d. samples is to cycle over a randomly permuted training set~\cite{online2}. The convergence of the overall algorithm is proved empirically and theoretically \cite{online}. We do not elaborate the proofs as the main contribution is not in the proof, and interested readers are encouraged to refer this paper~\cite{online}, where the proofs have been already available.

\begin{algorithm}[!ht]
\caption{The online discriminative algorithm for dictionary learning} \label{alg:1}
\textbf{Input}: $(\x,\y) \in \RB^n \times \RB^q \sim p(\x,\y)$ (random variables and a method to draw i.i.d samples of $p$), $\lambda_0 \in \RB$ (regularization parameters), $L \in \RB$ (sparsity factor), $T$ (number of iterations).

\textbf{Output}: Dictionary $\D$ and classifier parameter $\W$.

\begin{algorithmic}[1]
    \STATE Initialize the dictionary $\D$ and classifier $\W$.
    \STATE Set $\M_0 \in \RB^{k \times k} = 0$, and $\N_0 \in \RB^{{(n+q)} \times k} = 0$
    \WHILE{stop criterion is not reached or $t = 1$ to $T$}
        \STATE Draw $(\x_t,\y_t)$ from $p(\x,\y)$
        \STATE Sparse coding: compute $\alp_t$ via solving the following optimization problem:
        \begin{displaymath}
        \alp_t = \mathop{\arg \min}\limits_{\alp \in \RB^k} \frac{1}{2}\mathop{\left\|\x_t-\D\alp\right\|}\nolimits_2^2,\textrm { \emph{s.t.} }\mathop{\left\|\alp\right\|}\nolimits_0 \le L
        \end{displaymath}
        \STATE $\M_t = \M_{t-1}+\alp_t{\alp_t}^T$.
        \STATE $\N_t = \N_{t-1}+\widetilde{\x_t}{\alp_t}^T$.
        \STATE Update the parameters $\D$ and $\W$ by a block-coordinate descent method in Algorithm~\ref{alg:2}.
        \STATE Normalize the columns of $\D$ such that the $\ell^2$ norm of each column equals to 1.
    \ENDWHILE
    \STATE Return {$\D$ and $\W$}
\end{algorithmic}
\end{algorithm}

\begin{algorithm}[!ht]
\caption{Dictionary and classifier parameter update} \label{alg:2}
\textbf{Input}: $\mathop{\widetilde{\D}}\nolimits_{t-1} \in \RB^{{(n+q)} \times k}$, $\M_t \in \RB^{k \times k}$, $\N_t \in \RB^{{(n+q)} \times k}$.

\textbf{Output}: $\D_t$ and $\W_t$.

\begin{algorithmic}[1]
    \REPEAT
        \FOR{$l = 1$ to $k$}
            \STATE Update the $l$-th columns of $\mathop{\widetilde{\D}}\nolimits_t$ using
            \begin{displaymath}
            \mathop{\widetilde{\bf{d}}}\nolimits_l = \frac{1}{m_{ll}^{t-1}}(\n_l^{t-1}-\mathop{\widetilde{\D}}\nolimits_{t-1}\m_l^{t-1})+\mathop{\widetilde{\bf{d}}}\nolimits_l^{t-1}
            \end{displaymath}
            where the superscript $t-1$ denotes the results from the $(t-1)$-th iteration.
            \STATE Separate $\mathop{\widetilde{\d}}\nolimits_l$ as $\d_l$ and $\w_l$.
            \STATE Update $\d_l$ and $\w_l$ using
            \begin{displaymath}
            \begin{array}{c}
            \d_l = \d_l/{\left\|\d_l\right\|}_2 \\
            \w_l = \w_l/{\left\|\d_l\right\|}_2
            \end{array}
            \end{displaymath}
        \ENDFOR
    \UNTIL \textbf{convergence}
    \STATE Return {$\D_t$ and $\W_t$}
\end{algorithmic}
\end{algorithm}

\textbf{Initialization.} The initialization of dictionary $\D$ and classifier $\W$ plays an important role in our proposed method. It may lead to poor performances if they are not well initialized. One can use patches randomly selected from the training data and zero matrix to initialize $\D$ and $\W$ respectively. In practice, our experiments show that using the classical reconstructive dictionary as our initial dictionary $\D$ always lead to better performances than that of original patches from the training data. 
Using this initial dictionary $\D$, the classifier $\W$ can be initialized via solving the optimization problem (\ref{equa:4}).

\textbf{Mini-batch strategy.} The convergence speed of our algorithm can be improved with a mini-batch strategy, which is widely used in stochastic gradient descent algorithms. The mini-batch strategy draws more than one samples (denote the number of samples as $\kappa$) at each iteration instead of a signal one. This is inspired by the fact the runtime for solving $\kappa$ $\ell^0$ pseudo norm optimization problem (\ref{equa:1}) with dictionary $\D$ can be greatly shorten using Batch-OMP algorithm~\cite{Batchomp} with precomputation of matrix $\D^T\D$.

\section{Experiment}
\label{sec:exp}

In this section~\footnote{Our propose ODDL method is an online approach, therefore testing on a large scale database is a requisite to evaluate the performance. However, the large-scale database evaluation is under way and we plan to report it along with one of our future work.}, we demonstrate the performance of our proposed ODDL method in two image classification tasks, handwritten digit recognition and face recognition.
Before presenting the experiments, we first discuss the choices of three important parameters in our algorithm.

\subsection{Choices of Parameters}

\textbf{Parameter $L$.} As introduced in the previous section, in our algorithm we choose the $\ell^0$ pseudo norm optimization problem as our sparse coding problem and use the Orthogonal Matching Pursuit (OMP) algorithm to find the approximative solutions. The sparsity prior $L$ controls the nonzero elements of the sparse coefficients in our algorithm. Our experiments have shown that handwritten digit images and face images can be represented well when $L$ are $5$ and $15$ respectively.

\textbf{Parameter $\lambda_0$.} $\lambda_0$ is the parameter controlling the trade-off between the reconstructive and discriminative power in our method. $\lambda_0$ of large values will pay most attention to the reconstructive error, while small $\lambda_0$ would enhance the discriminative power at the cost of losing the representation ability. Thus, the value of $\lambda_0$ plays an important role for balancing representation and classification. In practice, the value $\lambda_0 = 1$ has given good performances in our experiments.

\textbf{Parameter $T$.} In our method, we cycle over a randomly permuted training set which is a common technique in online algorithms to obtain i.i.d. samples for experiments. We have observed that when $T$ is such a value that the whole training set is cycled one round the experimental results are always good.

\subsection{Handwritten Digit Recognition}

In this section we present experiments on the MNIST~\cite{MNIST} and USPS~\cite{USPS} handwritten digit datasets. MNIST contains a total number of $70000$ images of size $28 \times 28$, in which there are $60000$ images for training and $10000$ images for testing. USPS contains $7291$ training images and $2007$ testing images of size $16 \times 16$.

All the digit images are vectored and normalized to have zero mean and unit $\ell^2$ norm. Using these two datasets, we test four methods: our proposed \textbf{ODDL} method, ksvd method with a linear classifier, dubbed \textbf{ksvd-linear}, online reconstructive dictionary learning method with a linear classifier, dubbed \textbf{online-rec-linear}, and \textbf{dksvd} (referred to~\cite{Zhang10}) method. In \textbf{ODDL} and \textbf{dksvd} methods, we learn a signal dictionary $\D$ with $960$ atoms, corresponding to roughly $96$ atoms each class, and a signal classifier. While for \textbf{ksvd-linear} and \textbf{online-rec-linear} methods, first $10$ independent dictionaries each with $96$ atoms are learned, one for each class. Then, we adopt the one-vs-all strategy~\cite{1-vs-all} for learning classifiers. For class $i$, the one-vs-all strategy uses all samples from class $i$ as the positive samples and samples from the other classes as the negative samples to train the classifier of class $i$.

\begin{figure}
\centering
    \begin{minipage}[a]{0.9\linewidth}
    \centering
    \includegraphics[width=0.8\linewidth]{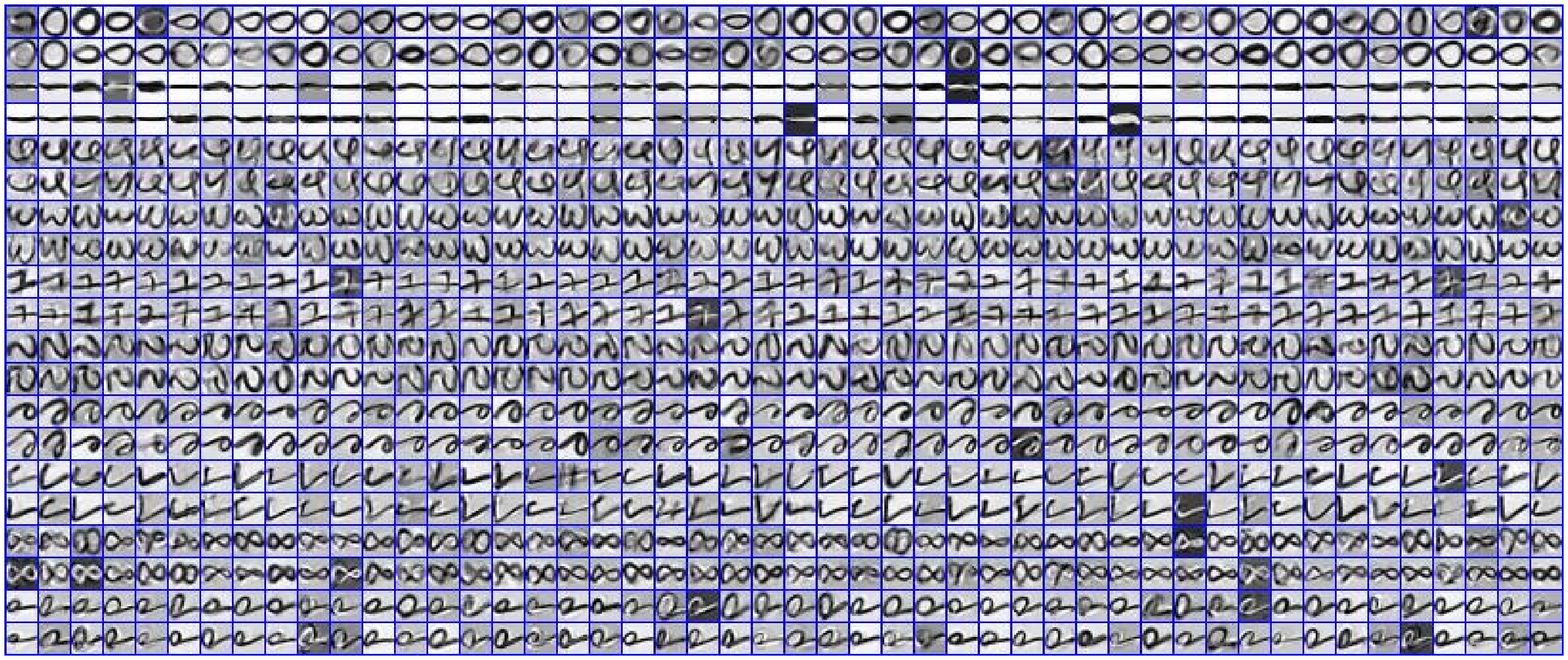}
    \vspace{1mm}
    \end{minipage}
    \begin{minipage}[a]{0.9\linewidth}
    \centering
    \includegraphics[width=0.8\linewidth]{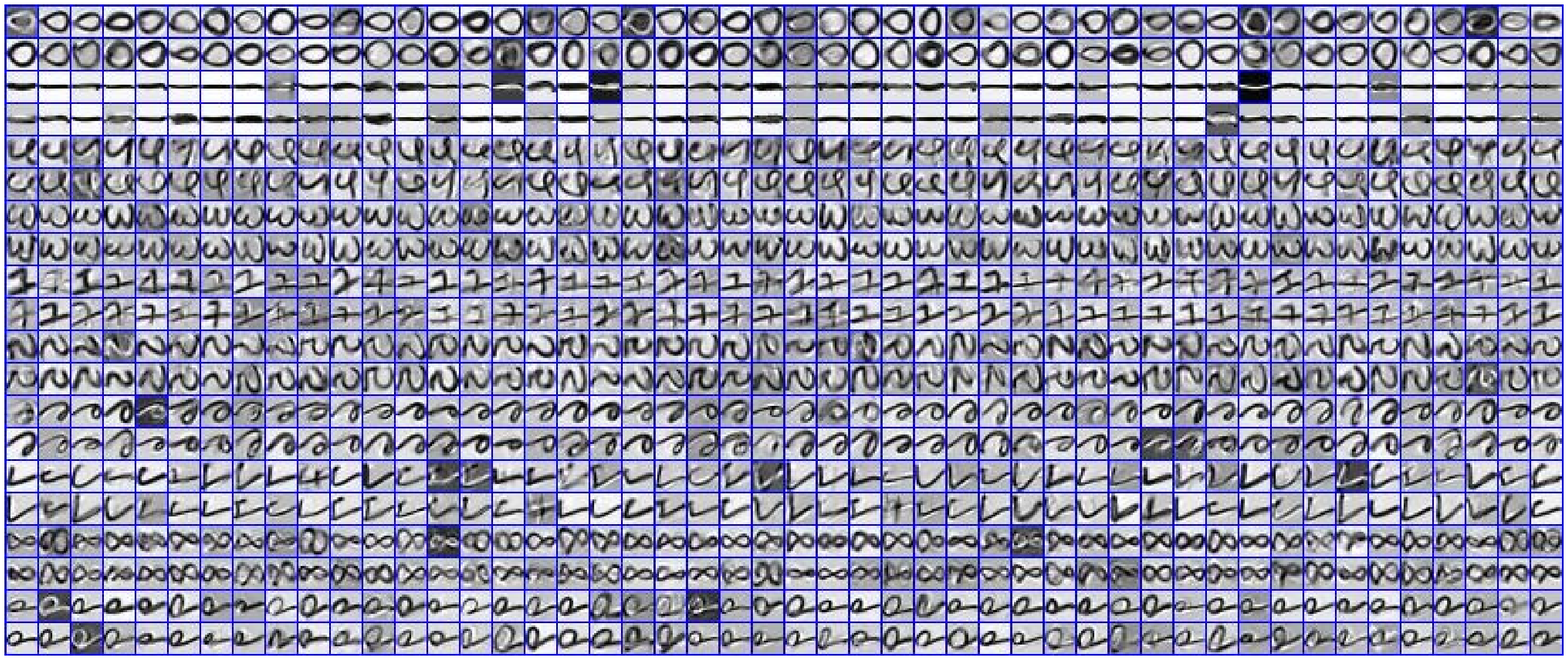}
    \vspace{1mm}
    \end{minipage}
\caption{Above: the learned dictionary in a reconstructive manner. Below: the learned dictionary by our ODDL method. 
}
\label{fig:2}
\end{figure}

The average error rates of four testing methods on MNIST and USPS are shown in Table~\ref{tab:1}. From the results, we can see that learning dictionaries in a discriminative way lead to better performance than those learned in a reconstructive way when adapted to classification task. When compared with those methods which use more sophisticated classifier models such as linear and bilinear logistic loss functions, our proposed method does not perform better. We believe that one of the main reasons is due to the simplicity of our linear classifier model. Our proposed method provides a new strategy for online discriminative dictionary learning, and the great strength is that in our framework the dictionary and classifier can be updated jointly, markedly different from the strategy of dictionary and classifier training in most existing methods. Figure~\ref{fig:2} shows dictionaries of the USPS dataset, which are learned via \textbf{ksvd-linear} and \textbf{ODDL} methods respectively.

\begin{table}
\begin{center}
\caption{Average error rates of testing methods for the MNIST and USPS datasets.}
\vspace{2mm}
\label{tab:1}
\begin{tabular}{|c|c|c|}
\hline
Method & MNIST & USPS\\
\hline
ODDL  & \textbf{3.58} & \textbf{5.35}\\
\hline
ksvd-linear  & 5.07 & 7.12\\
\hline
online-rec-linear & 5.32 & 7.35\\
\hline
dksvd  & 4.58 & 6.53\\
\hline
\end{tabular}
\end{center}
\end{table}

In addition, we also compare the runtime of our \textbf{ODDL} method and the \textbf{ksvd-linear} method for dictionary and classifier training. We take the total time for learning dictionaries and classifiers for all classes, then computed the average runtime via dividing it by the number of classes. 
The results are shown in Table~\ref{tab:2}. 
From Table~\ref{tab:2}, we can see our proposed ODDL can largely shorten the runtime for dictionary and classifier learning compared with the ksvd-linear method with the same dictionary size.

\begin{table}
\begin{center}
\caption{Average runtime (\emph{s}) for training stage using our proposed method and the ksvd-linear method.}
\vspace{2mm}
\label{tab:2}
\begin{tabular}{|c|c|c|}
\hline
Method & MNIST & USPS\\
\hline
ODDL & \textbf{156} & \textbf{23}\\
\hline
ksvd-linear & 583 & 62\\
\hline
\end{tabular}
\end{center}
\end{table}

To study the role of the dictionary size in our method, we proceed another set of experiments. We learn dictionaries from the training set with different sizes $k$ in $\{160,320,640,960,1280,2560\}$, and record the performances of these dictionaries on the testing set. The results are shown in Table~\ref{tab:3}. We observe that the dictionary size plays an important role in classification task. If $k$ is too small, information in learned dictionaries is not sufficient for discriminative. When $k$ is too big, learned dictionaries contain too much redundant information which may influence discrimination.

\begin{table}
\begin{center}
\caption{Average error rates of our proposed method for the MNIST and USPS datasets with different values of dictionary size $k$. }
\vspace{2mm}
\label{tab:3}
\begin{tabular}{|c|c|c|c|c|c|c|}
\hline
k & 160 & 320 & 640 & 960 & 1280 & 2560\\
\hline
MNIST  & 5.49 & 4.76 & 4.02 & \textbf{3.58} & 3.92 & 4.38\\
\hline
USPS  & 7.63 & 6.43 & 5.78 & \textbf{5.35} & 5.69 & 6.24\\
\hline
\end{tabular}
\end{center}
\end{table}

\subsection{Extended YaleB Face Recognition}

\begin{figure}
\centering
    \begin{minipage}[a]{.3\linewidth}
    \centering
    \includegraphics[width=.9\linewidth]{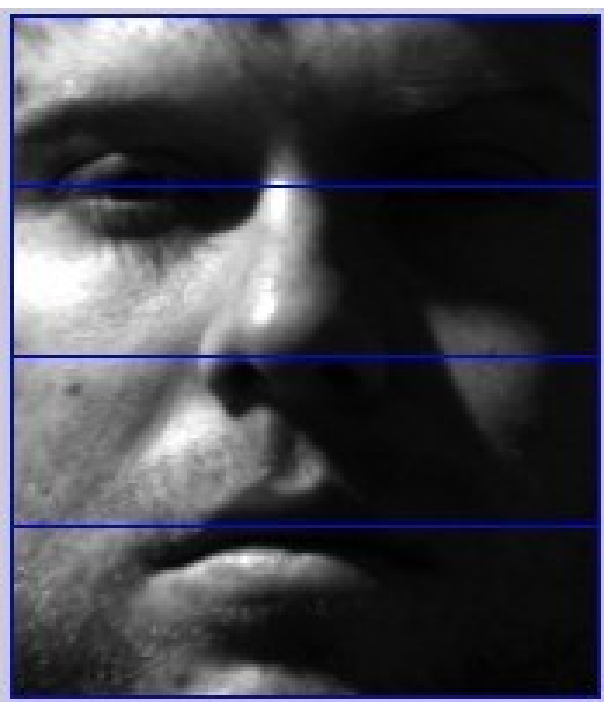}
    \vspace{1mm}
    \end{minipage}
    \begin{minipage}[a]{.3\linewidth}
    \centering
    \includegraphics[width=.9\linewidth]{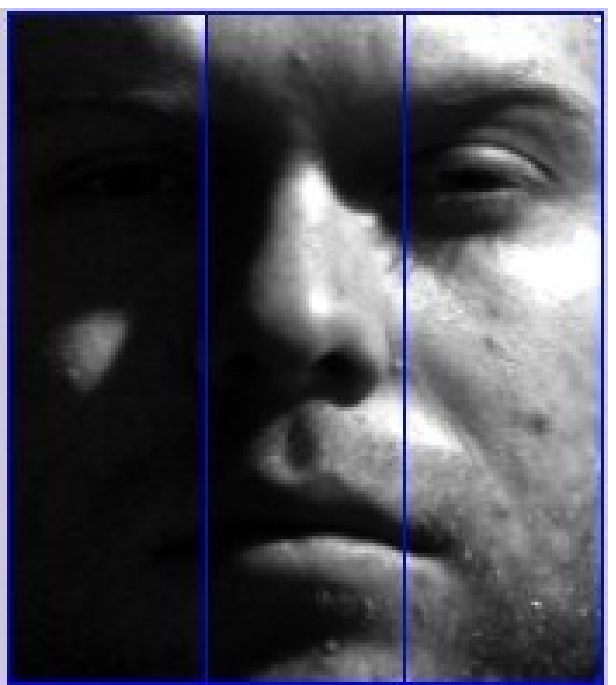}
    \vspace{1mm}
    \end{minipage}
\caption{Original patches drawn from face images.}
\label{fig:3}
\end{figure}

The Extended YaleB face dataset~\cite{yaleB} consists of $2414$ near frontal face images of $38$ individuals. These images are taken with different poses and under different illumination conditions. 
We randomly divide the dataset into two parts, and each part contains approximate $26$ samples. One is used for learning the dictionary and classifier, while the other is used as the testing set. 
Before presenting our experiments, we need some pre-processing steps. As known, the most important features in face recognition are eyebrows, eyes, nose, mouse, and chin. Using this information, we divide each face image into four non-overlapping patches from top to bottom, and into three non-overlapping patches from left to right. 
Figure~\ref{fig:3} shows such patches.
We can observe that each patch contains at least one feature. After doing this, for each person we have seven classes of patches. Then we vector all the patches and normalized them to have unit $\ell^2$ norm. In our experiments, seven dictionaries with $228$ atoms and seven classifiers are learned, corresponding to seven patch class.

\begin{figure}
\centering
    \begin{minipage}[a]{0.9\linewidth}
    \centering
    \includegraphics[width=0.8\linewidth]{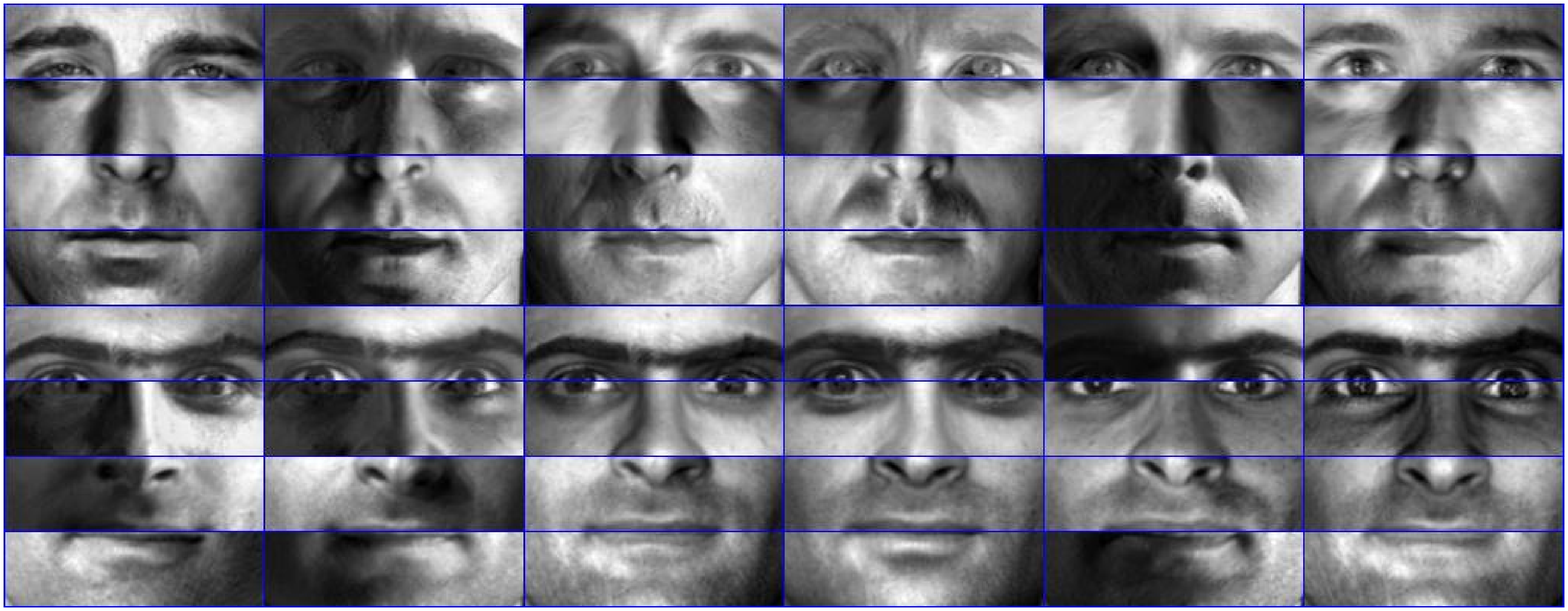}
    \vspace{1mm}
    \end{minipage}
    \begin{minipage}[a]{0.9\linewidth}
    \centering
    \includegraphics[width=0.8\linewidth]{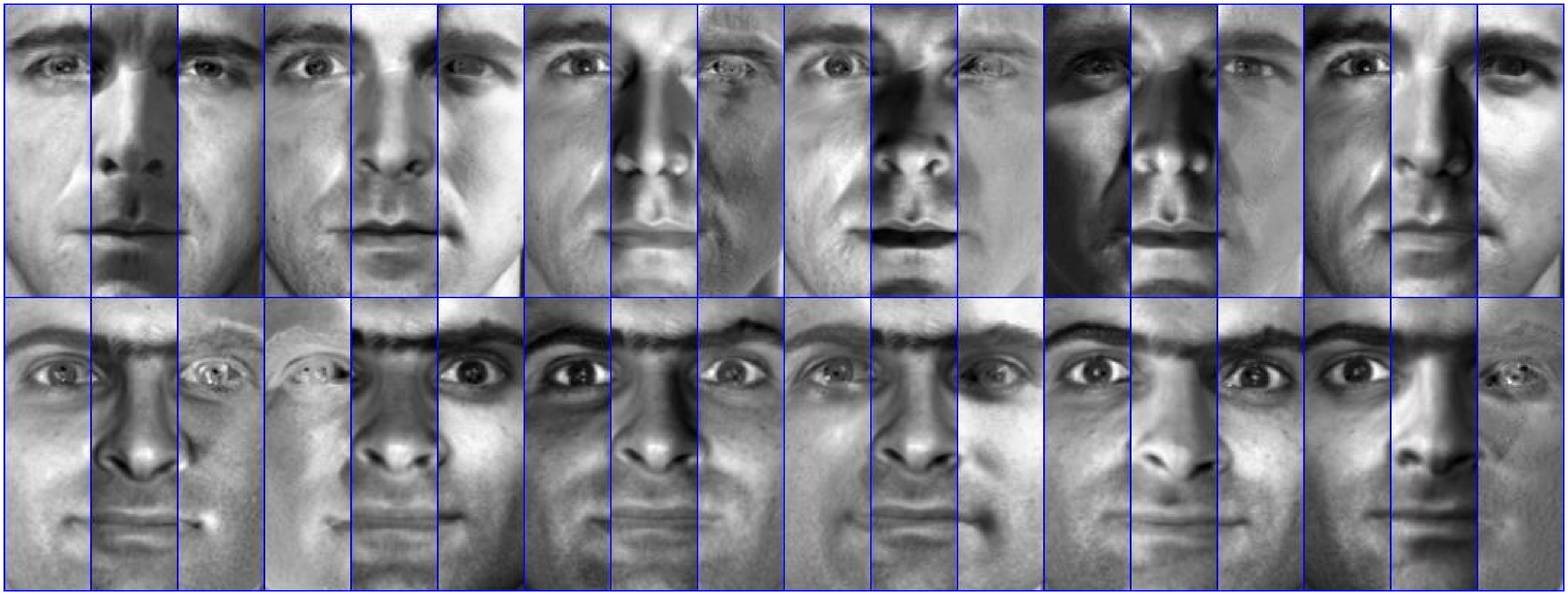}
    \vspace{1mm}
    \end{minipage}
\caption{Dictionaries learned via our proposed ODDL. Here we manually rearrange the seven learned dictionaries to two big dictionaries.}
\label{fig:4}
\end{figure}

For comparison, we also test our proposed method with \textbf{ksvd-linear}, \textbf{online-linear}, and \textbf{dksvd} methods. 
The results are demonstrated in Table~\ref{tab:4}. It is easy to see that discriminative dictionary performs better than reconstructive dictionaries.
Figure~\ref{fig:4} plots the dictionaries learned by our ODDL method for two individuals.

\begin{table}
\begin{center}
\caption{Average error rates of testing methods for the Extended YaleB face dataset.}
\vspace{2mm}
\label{tab:4}
\begin{tabular}{|c|c|c|c|}
\hline
ODDL & ksvd-linear & online-linear & dksvd\\
\hline
\textbf{1.09}  & 2.03 & 2.24 & 1.76\\
\hline
\end{tabular}
\end{center}
\end{table}

As in the experiments with handwritten digit datasets, we also compared the average runtime of training stage of our proposed ODDL method and the \textbf{ksvd-linear} method. Table~\ref{tab:5} shows the final results. For the \textbf{ksvd-linear} method, the dictionary size is $6$ for each patch class of each person. For our proposed method, we test the average runtime of training stage when the dictionary sizes are $228$ and $456$ respectively. As expected, learning dictionaries with smaller size can shorten the runtime.

\begin{table}
\begin{center}
\caption{Average runtime (\emph{s}) for training stage using our proposed method and the ksvd-linear method.}
\vspace{2mm}
\label{tab:5}
\begin{tabular}{|c|c|c|}
\hline
ODDL (304) & ODDL (608) & ksvd-linear\\
\hline
3 & 4 & 10\\
\hline
\end{tabular}
\end{center}
\end{table}

\section{Conclusion and Future Work}
\label{sec:conclusion}
In this paper, we propose a novel framework for online discriminative dictionary learning (ODDL) for image classification task. 
By introducing a linear classifier into the conventional dictionary learning problem, the learned dictionary will capture the discriminative cues for classification along with representation powerfulness for reconstruction. 
We propose an online algorithm to solve this discriminative problem. 
Unlike other algorithms which find the dictionary and classifier alternately via solving the sub-problems iteratively, our algorithm directly finds them jointly. 
The experimental results on MNIST and USPS handwritten digit datasets and the Extended yaleB face dataset demonstrate that our method is very competitive when applied to image classification task with large-scale training set. 
More experiments need to be done to better demonstrate the performances of our proposed methods for image classification  in the future.

\subsection*{Acknowledgements}
This work is supported by by 973 Program (Project No.2010CB327905) and Natural Science Foundations of China (No.61071218).

\appendix
\section{Appendix}

To obtain (\ref{equa:12}), denote $f(\D,\W)$ as the function to minimize in (\ref{equa:11}), then a bit of algebra gives
\begin{equation}
\begin{array}{rcl}
f(\D,\W) & = & \mathop{\left\|{\widetilde{\X}}_t-\widetilde{\D}\A_t\right\|}\nolimits_F^2\\
& = & Tr[{({\widetilde{\X}}_t-\widetilde{\D}\A_t)}^T({\widetilde{\X}}_t-\widetilde{\D}\A_t)]\\
& = & Tr({\widetilde{\D}}^T\widetilde{\D}\M_t)-2Tr({\widetilde{\D}}^T\N_t)+Tr({{\widetilde{\X}}_t}^T{\widetilde{\X}}_t)
\label{equa:15}
\end{array}
\end{equation}
where ${\widetilde{\X}}_t = [{\widetilde{\x}}_1,...,{\widetilde{\x}}_t]$, and $\A_t = [\alp_1,...,\alp_t]$. Since the last term of the final formulation is irrespective of $\D$ and $\W$, dropping it then we can obtain (\ref{equa:12}).

In order to obtain the update of ${\widetilde{\bf{d}}}_j$, the $j$-th column of $\widetilde{\D}$, a block-coordinate descent method is used. Denote the objective function in (\ref{equa:12}) as $f(\widetilde{\D})$, then using some algebraic transformations we obtain
\begin{equation}
f(\widetilde{\D}) = \mathop{\sum}\limits_i{{\widetilde{\bf{d}}}_i}^T\mathop{\sum}\limits_l{\widetilde{\bf{d}}}_lm_{il}-2\mathop{\sum}\limits_i{{\widetilde{\bf{d}}}_i}^T\n_i
\label{equa:16}
\end{equation}
Now consider only the terms associated with ${\widetilde{\bf{d}}}_j$, which we denote as $f({\widetilde{\bf{d}}}_j)$
\begin{equation}
\begin{array}{rcl}
f({\widetilde{\bf{d}}}_j) & = & ({{\widetilde{\bf{d}}}_j}^T\mathop{\sum}\limits_{l \ne j}{\widetilde{\bf{d}}}_lm_{jl}+{{\widetilde{\bf{d}}}_j}^T{\widetilde{\bf{d}}}_jm_{jj}+\mathop{\sum}\limits_{i \ne j}{{\widetilde{\bf{d}}}_i}^T{\widetilde{\bf{d}}}_jm_{ij})-\\
& & 2{{\widetilde{\bf{d}}}_j}^T\n_j\\
& = & ({{\widetilde{\bf{d}}}_j}^T{\widetilde{\bf{d}}}_jm_{jj}+{{\widetilde{\bf{d}}}_j}^T\mathop{\sum}\limits_{l \ne j}{\widetilde{\bf{d}}}_lm_{jl}+\mathop{\sum}\limits_{i \ne j}{{\widetilde{\bf{d}}}_i}^T{\widetilde{\bf{d}}}_jm_{ji})-\\
& & 2{{\widetilde{\bf{d}}}_j}^T\n_j\\
& = & ({{\widetilde{\bf{d}}}_j}^T{\widetilde{\bf{d}}}_jm_{jj}+{{\widetilde{\bf{d}}}_j}^T\mathop{\sum}\limits_{l \ne j}{\widetilde{\bf{d}}}_lm_{jl}+{{\widetilde{\bf{d}}}_j}^T\mathop{\sum}\limits_{i \ne j}{\widetilde{\bf{d}}}_im_{ji})-\\
& & 2{{\widetilde{\bf{d}}}_j}^T\n_j\\
& = & {{\widetilde{\bf{d}}}_j}^T{\widetilde{\bf{d}}}_jm_{jj}+2{{\widetilde{\bf{d}}}_j}^T\mathop{\sum}\limits_{l \ne j}{\widetilde{\bf{d}}}_lm_{jl}-2{{\widetilde{\bf{d}}}_j}^T\n_j
\label{equa:17}
\end{array}
\end{equation}
Notice in above transformations we use an important information that the matrix $\M_t$ is symmetric. Computing the derivative of $f({\widetilde{\bf{d}}}_j)$ with respect to ${\widetilde{\bf{d}}}_j$ we have
\begin{equation}
\frac{\partial{f({\widetilde{\bf{d}}}_j)}}{\partial{{\widetilde{\bf{d}}}_j}} = 2m_{jj}{\widetilde{\bf{d}}}_j+2\mathop{\sum}\limits_{l \ne j}{\widetilde{\bf{d}}}_jm_{jl}-2\n_j
\label{equa:18}
\end{equation}
Thus setting the above derivative to 0, ${\widetilde{\bf{d}}}_j$ can be updated
\begin{equation}
\begin{array}{rcl}
{\widetilde{\bf{d}}}_j & = & \frac{\n_j-\mathop{\sum}\limits_{l \ne j}{\widetilde{\bf{d}}}_lm_{il}}{m_{jj}}\\
& = & \frac{\n_j-\mathop{\sum}\limits_{l}{\widetilde{\bf{d}}}_lm_{il}+{\widetilde{\bf{d}}}_jm_{jj}}{m_{jj}}\\
& = & \frac{1}{m_{jj}}(\n_j-\widetilde{\D}\m_j)+{\widetilde{\bf{d}}}_j
\label{equa:19}
\end{array}
\end{equation}

\bibliography{bib}
\bibliographystyle{ieee} 

\end{document}